\documentclass[10pt,twocolumn,letterpaper]{article}

\usepackage{cvpr}              % To produce the CAMERA-READY version
\usepackage{graphicx}
\usepackage{amsmath}
\usepackage{amssymb}
\usepackage{booktabs}
\usepackage{bm}
\usepackage{graphicx}
\usepackage{amsmath}
\usepackage{amssymb}
\usepackage{booktabs}
\usepackage{bm}
\usepackage{times}
\usepackage{epsfig}
\usepackage{graphicx}
\usepackage{amsmath}
\usepackage{amssymb}
\usepackage{color,soul}
\usepackage{multirow}
\usepackage{enumitem}
\usepackage[normalem]{ulem}
\usepackage{algorithm}
\usepackage{xcolor}
\usepackage{listings}
\usepackage[font=small]{caption}
\usepackage[font=footnotesize]{subcaption}
\usepackage{stmaryrd}
\usepackage{colortbl}
\usepackage{booktabs} 
\definecolor{mygray}{gray}{.9}
\usepackage[pagebackref,breaklinks,colorlinks]{hyperref}

\usepackage[capitalize]{cleveref}
\crefname{section}{Sec.}{Secs.}
\Crefname{section}{Section}{Sections}
\Crefname{table}{Table}{Tables}
\crefname{table}{Tab.}{Tabs.}

\def\cvprPaperID{8171} % *** Enter the CVPR Paper ID here
\def\confName{CVPR}
\def\confYear{2023}

\begin{document}

%%%%%%%%% TITLE - PLEASE UPDATE
\title{Person Image Synthesis via Denoising Diffusion Model}

\author{Ankan Kumar Bhunia\textsuperscript{1} \hspace{.1cm} Salman Khan\textsuperscript{1,2} \hspace{.1cm} Hisham Cholakkal\textsuperscript{1} \hspace{.1cm} Rao Muhammad Anwer\textsuperscript{1,4}\\Jorma Laaksonen\textsuperscript{4} \hspace{.1cm} Mubarak Shah\textsuperscript{5} \hspace{.1cm} Fahad Shahbaz Khan\textsuperscript{1,3} \\
\textsuperscript{1}Mohamed bin Zayed University of AI, UAE \hspace{.1cm} \textsuperscript{2}Australian National University, Australia
\hspace{.1cm}  \\ \textsuperscript{3}Link{\"o}ping University, Sweden  \hspace{.1cm} \textsuperscript{4}Aalto University, Finland  \hspace{.1cm} \textsuperscript{5}University of Central Florida, USA\\
%{\tt\small \textsuperscript{1}ankan.bhunia@mbzuai.ac.ae }
}
\maketitle

%%%%%%%%% ABSTRACT
\begin{abstract}
The pose-guided person image generation task requires synthesizing photorealistic images of humans in arbitrary poses. The existing approaches use generative adversarial networks that do not necessarily maintain realistic textures or need dense correspondences that struggle to handle complex deformations and severe occlusions. In this work, we show how denoising diffusion models can be applied for high-fidelity person image synthesis with strong sample diversity and enhanced mode coverage of the learnt data distribution. Our proposed Person Image Diffusion Model (PIDM) disintegrates the complex transfer problem into a series of simpler forward-backward denoising steps. This helps in learning plausible source-to-target transformation trajectories that result in faithful textures and undistorted appearance details. We introduce a `{texture diffusion module}' based on cross-attention to accurately model the correspondences between appearance and pose information available in source and target images. Further, we propose `{disentangled classifier-free guidance}' to ensure close resemblance between the conditional inputs and the synthesized output in terms of both pose and appearance information. Our extensive results on two large-scale benchmarks and a user study demonstrate the photorealism of our proposed approach under challenging scenarios. We also show how our generated images can help in downstream tasks. Our code and models will be publicly released. 
\end{abstract}

\begin{figure}[t!]
\begin{center}
   \includegraphics[width=1\linewidth]{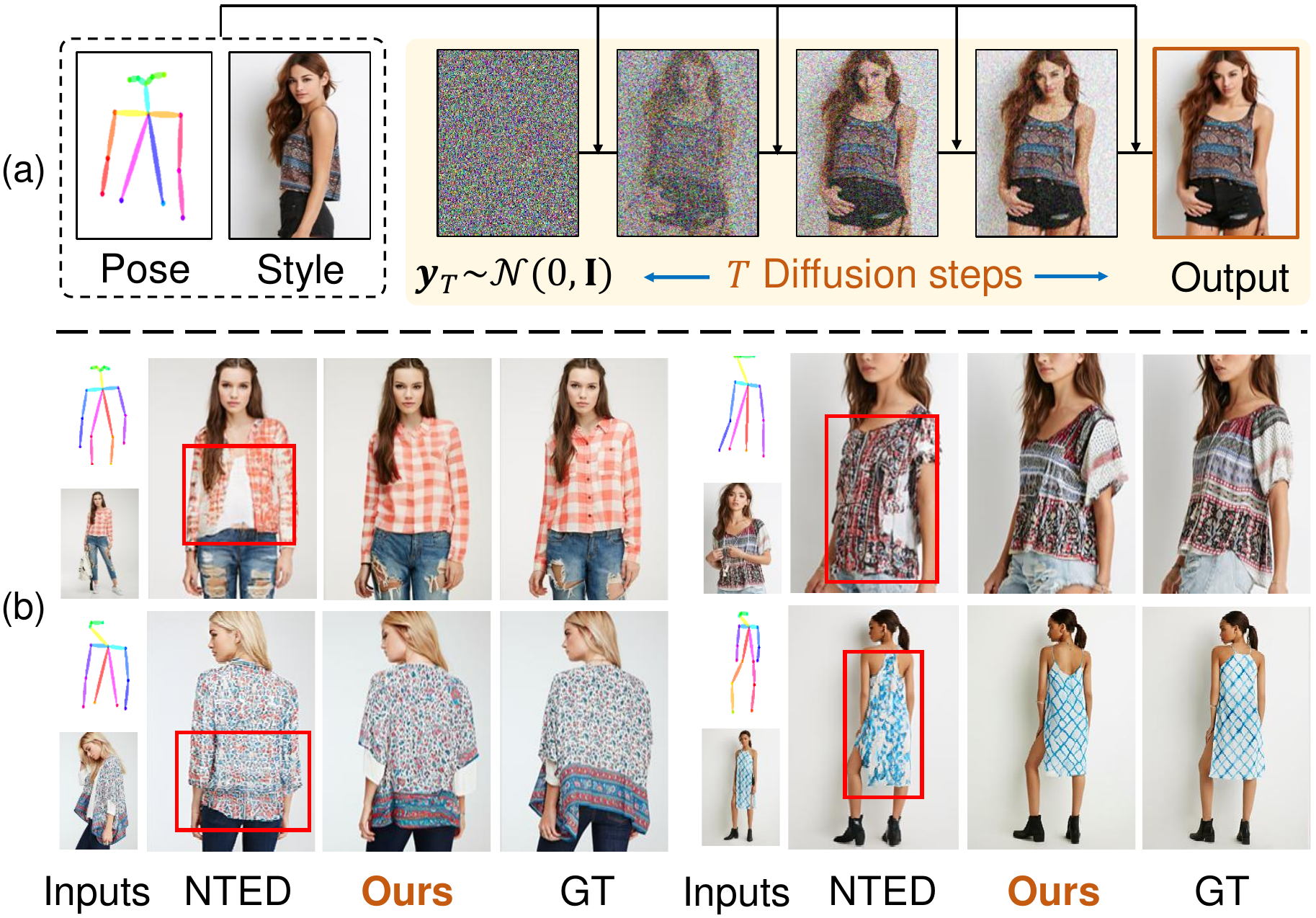}\vspace{-0.4cm}
\end{center}
\caption{\textbf{(a)} Our proposed PIDM is a denoising diffusion model where the generative path is conditioned on the pose and style. PIDM breaks down the problem into a series of forward-backward diffusion steps to learn the plausible transfer trajectories. \textbf{(b)} Comparison of PIDM with the recently introduced NTED~\cite{ren2022neural}. PIDM accurately retains the appearance of the source style image while also producing images that are more natural and sharper while NTED struggles to adequately preserve the source appearance in complex scenarios (marked in red boxes).
}
\label{fig:intro}
 \vspace{-0.1cm}
\end{figure}
%%%%%%%%% BODY TEXT
\section{Introduction}
\label{sec:intro}
The Pose-guided person image synthesis task~\cite{zhu2019progressive,ren2020deep, zhang2021pise} aims to render a person’s image with a desired pose and appearance. Specifically, the appearance is defined by a given source image and the pose by a set of keypoints. Having control over the synthesized person images in terms of pose and style is an important requisite for applications such as e-commerce, virtual reality, metaverse and content generation for the entertainment industry. Furthermore, the generated images can be used to improve performance on downstream tasks such as person re-identification~\cite{zhu2019progressive}. The challenge is to generate photorealistic outputs tightly conforming with the given pose and appearance information.

In the literature, person synthesis problem is generally tackled using Generative Adversarial Networks (GAN)~\cite{goodfellow2020generative} which try to generate a person in a desired pose using a single forward pass. However, preserving a coherent structural, appearance and global body composition in the new pose is a challenging task to achieve in one shot. The resulting outputs commonly experience deformed textures and unrealistic body shapes, especially when synthesizing occluded body parts (see Fig.~\ref{fig:intro}). Further, GANs are prone to unstable training behaviour due to adversarial min-max objective and lead to limited diversity in the generated samples. Similarly, Variational Autoencoder~\cite{kingma2013auto} based solutions have been explored that are relatively stable, but suffer from blurry details and offer low-quality outputs than GANs due to their dependence on a surrogate loss for optimization.

In this work, we frame the person synthesis problem as a series of diffusion steps that progressively transfer a person in the source image to the target pose. Diffusion models~\cite{ho2020denoising} are motivated from non-equilibrium thermodynamics that define a Markov chain of slowly adding noise to the input samples (forward pass) and then reconstructing the desired samples from noise (reverse pass). In this manner, rather than modeling the complex transfer characteristics in a single go, our proposed person synthesis approach PIDM breaks down the problem into a series of forward-backward diffusion steps to learn the plausible transfer trajectories. Our approach can model the intricate interplay of the person’s pose and appearance, offers higher diversity and leads to photorealistic results without texture deformations (see Fig.~\ref{fig:intro}). In contrast to existing approaches that deal with major pose shifts by requiring parser maps denoting human body parts~\cite{men2020controllable, zhang2021pise, zhou2022cross}, or dense 3D correspondences~\cite{li2019dense, liu2019liquid} to fit human body topology by warping the textures, our approach can learn to generate realistic and authentic images without such detailed annotations.

Our major contributions are as follows:\vspace{-0.5em}
\begin{itemize}\setlength{\itemsep}{0em}
\item We develop the first \emph{diffusion-based approach} for pose-guided person synthesis task which can work under challenging pose transformations while preserving appearance, texture and global shape characteristics. 
\item To effectively model the complex interplay between appearance and pose information, we propose a \emph{texture diffusion module}. This module exploits the correspondences between source and target appearance and pose details, hereby obtaining artefact free images. 
\item In the sampling procedure, we introduce \emph{disentangled classifier-free guidance} to tightly align the output image style and pose with the source image appearance and target pose, respectively. It ensures close resemblance between the conditions that are input to the generative model and the generated output.
\item Our results on DeepFashion~\cite{liu2016deepfashion} and Market-1501~\cite{zheng2015scalable} benchmarks set new state of the art. We also report a user study to evaluate the qualitative features of generated images. Finally, we demonstrate that synthesized images can be used to improve performance in downstream tasks \eg, person re-identification. 
\end{itemize}

\section{Related Work}
\textbf{Pose-guided Person Image Synthesis:}
The problem of human pose transfer has been studied extensively during the recent years, especially with the unprecedented success of GAN-based models~\cite{mirza2014conditional} for conditional image synthesis. An early attempt~\cite{ma2017pose} proposes a coarse-to-fine approach to first generate a rough image with the target pose and then refine the results adversarially. The method simply concatenates the source image, the source pose, and the target pose as inputs to obtain the target image, which leads to feature misalignment. To address this issue, Essner \etal~\cite{esser2018variational} attempt to disentangle the appearance and the pose of person images using VAE-based design and a UNet based skip-connection architecture. Siarohin \etal~\cite{siarohin2018deformable} improve the model by introducing deformable skip connections to spatially transform the textures, which decomposes the overall deformation by a set of local affine transformations. Subsequently, some works~\cite{li2019dense, liu2019liquid, ren2020deep} use flow-based deformation to transform the source information to improve pose alignment. Ren \etal~\cite{ren2020deep} propose GFLA that obtains the global flow fields and occlusion mask, which are used to warp local patches of the source image to match the required pose. Another group of works~\cite{li2019dense, liu2019liquid} use geometric models that fit a 3D mesh human model onto the 2D image, and subsequently predict the 3D flow, which finally warps the source appearance. On the other hand, without any deformation operation, Zhu \etal~\cite{zhu2019progressive} propose to progressively transform the source image by a sequence of transfer blocks. However, useful information can be lost during multiple transfers, which may result in blurry details. ADGAN~\cite{men2020controllable} uses a texture encoder to extract style vectors for human body parts %within each semantic 
and gives them to several AdaIN residual blocks to synthesize the final image. Methods such as PISE~\cite{zhang2021pise}, SPGnet~\cite{lv2021learning} and CASD~\cite{zhou2022cross} make use of parsing maps to generate the final image. CoCosNet~\cite{zhang2020cross, Zhou_2021_CVPR} extracts dense correspondences between cross-domain images with attention-based operation. Recently, Ren \etal~\cite{ren2022neural} propose a framework NTED based on neural texture extraction and distribution operation, which achieves superior results. 

\textbf{Diffusion Models:}
The existing GAN-based approaches attempt to directly transfer the style of the source image to a given target pose, which requires the architecture to model complex transformation of pose. In this work, we present a diffusion-based framework named PIDM that breaks the pose transformation process into several conditional denoising diffusion steps, in which each step is relatively simple to model. Diffusion models~\cite{ho2020denoising} are recently proposed generative models that can synthesize high-quality images. After success in unconditional generation, these models are extended to work in conditional generation settings, demonstrating competitive or even better performance than GANs. For class-conditioned generation, Dhariwal \etal~\cite{NEURIPS2021_49ad23d1} introduce classifier-guided diffusion, which is later adapted by GLIDE~\cite{nichol2021glide} to enable conditioning over CLIP textual representations. Recently, Ho \etal~\cite{ho2022classifier} propose a Classifier-Free Guidance approach that enables conditioning without requiring pretraining of the classifiers. In this work, we develop the first \textit{diffusion-based} approach for pose-guided person synthesis task. We also introduce \textit{disentangled
classifier-free guidance} to tightly align the output image style and pose with the source image appearance
and target pose, respectively. 

\begin{figure*}[t!]
\begin{center}
   \includegraphics[width=1\textwidth]{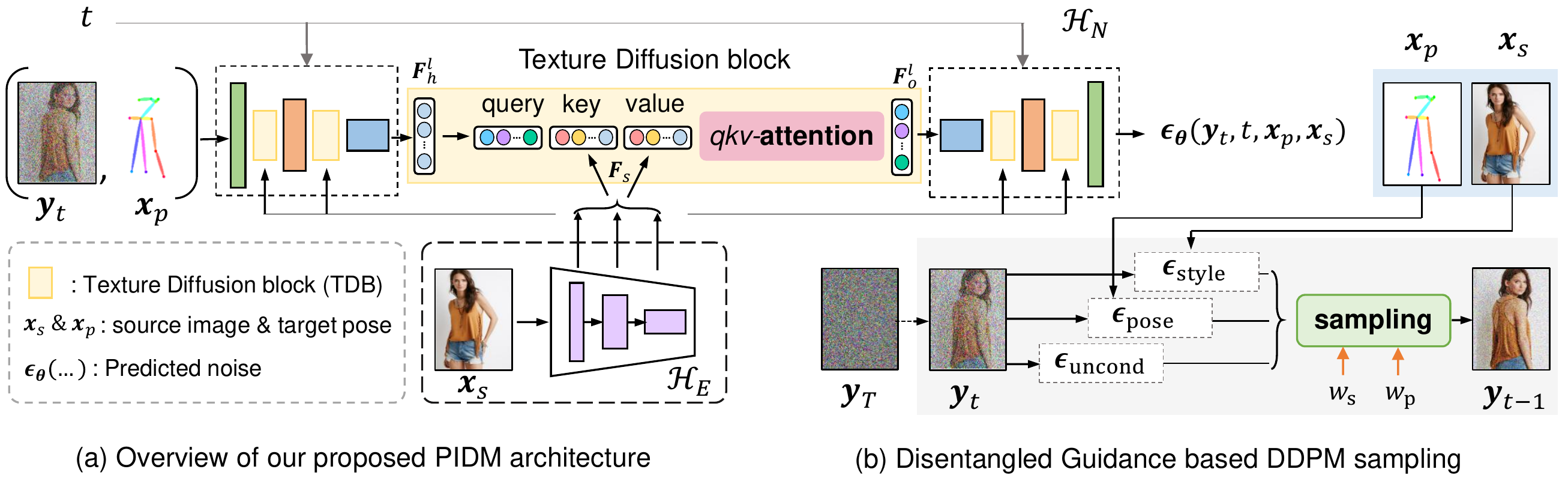}\vspace{-0.5cm}
\end{center}
\caption{(a) The proposed PIDM framework is a UNet-based network composed of a noise prediction module $\mathcal{H}_N$ and a texture encoder $\mathcal{H}_E$. The encoder $\mathcal{H}_E$ encodes the texture patterns of the source image $\bm{x}_s$. To obtain multi-scale features, we derive output from the different layers of $\mathcal{H}_E$ resulting in a stacked feature representation $\bm{F}_s$. To transfer rich multi-scale texture patterns from the source image distribution to the noise prediction module $\mathcal{H}_N$, we propose to use cross-attention based \textit{Texture diffusion blocks} (TDB) that are embedded in different layers of $\mathcal{H}_N$. This allows the network to fully exploit the correspondences between source and target appearance, thus resulting in distortion-free images. (b) During inference, to amplify the conditional signal of $\bm{x}_s$ and $\bm{x}_p$ in the sampled images, we adapt the Classifier-free guidance~\cite{ho2022classifier} in our sampling technique to achieve \textit{disentangled guidance}.
}
\label{fig:main}
 \vspace{-0.4cm}
\end{figure*}

\section{Proposed Method}
\textbf{Motivation:} The existing pose-guided person synthesis methods~\cite{ren2020deep, men2020controllable, zhang2021pise, lv2021learning, zhang2020cross, zhou2022cross, ren2022neural} rely on GAN-based frameworks where the model attempts to directly transfer the style of the source image into a given target pose in a single forward pass. It is quite challenging to directly capture the complex structure of the spatial transformation, therefore current CNN-based architectures often struggle to transfer the intricate details of the cloth texture patterns. As a result, the existing methods yield noticeable artifacts, which become more evident when the generator needs to infer %unobserved 
occluded body regions from the given source image.  

Motivated by these observations, we advocate that instead of learning the complex structure directly in a single step, deriving the final image using successive intermediate transfer steps can make the learning task simpler. 
%it would be easier to derive the final image by transferring it through a sequence of intermediate images.
To enable the above progressive transformation scheme, we introduce \textit{Person Image Diffusion Model} or PIDM, a diffusion-based~\cite{ho2020denoising} person image synthesis framework that breaks down the generation process into several conditional denoising diffusion steps, each step being relatively simple to model. A single step in the diffusion process can be approximated by a simple isotropic Gaussian distribution. We observe that our diffusion-based texture transfer technique PIDM can bring the following benefits: \textbf{(1)} \textit{High-quality synthesis}: Compared to previous GAN-based approaches, PIDM generates photo-realistic results when dealing with complex cloth-textures and extreme pose angles. \textbf{(2)} \textit{Stable training:} Existing GAN-based approaches use multiple loss objectives alongside adversarial losses, which are often difficult to balance, resulting in unstable training. In contrast, PIDM exhibits better training stability and mode coverage. Also, our model is less prone to hyperparameters. \textbf{(3)} \textit{Meaningful interpolation:} Our proposed PIDM allows us to achieve smooth and consistent linear interpolation in the latent space. \textbf{(4)} \textit{Flexibility:} The models from existing work are usually task dependent, requiring different models for various tasks (\eg, separate models for unconditional, pose-conditional, pose and style-conditional generation tasks). In contrast, in our case, a single model can be used to perform multiple tasks. Furthermore, PIDM inherits the flexibility and controllability of diffusion models that enable various downstream tasks (\eg, appearance control, see Fig.~\ref{fig:control}) using our model.

\textbf{Overall Framework:} Fig.~\ref{fig:main} shows the overview of the proposed generative model. Given a source image $\bm{x}_s$ and a target pose $\bm{x}_p$, our goal is to train a conditional diffusion model $p_{\theta}(\bm{y}|\bm{x}_s, \bm{x}_p)$ where the final output image $\bm{y}$ should not only satisfy the target pose matching requirement, but should also have the same style as in $\bm{x}_s$. 

The denoising network $\bm{\epsilon}_{\theta}$ in PIDM is a UNet-based design composed of a noise prediction module $\mathcal{H}_N$ and a texture encoder $\mathcal{H}_E$. The encoder $\mathcal{H}_E$ encodes the texture patterns of the source image $\bm{x}_s$. To obtain multi-scale features we derive output from the different layers of $\mathcal{H}_E$ resulting in stacked feature representation $\bm{F}_s=[\bm{f}_1, \bm{f}_2, ...,\bm{f}_m]$. To transfer rich multi-scale texture patterns from the source image distribution to the noise prediction module $\mathcal{H}_N$, we propose to use cross-attention based \textit{Texture diffusion blocks} (TDB) that are embedded in different layers of $\mathcal{H}_N$. This allows the network to fully exploit the correspondences between the source and target appearances, thus resulting in distortion-free images. During inference, to amplify the conditional signal of $\bm{x}_s$ and $\bm{x}_p$ in the sampled images, we adapt the classifier-free guidance~\cite{ho2022classifier} in our sampling technique to achieve \textit{disentangled guidance}. It not only improves the overall quality of the generation, but also ensures accurate transfer of texture patterns. %We use DDIM-based~\cite{song2020denoising} sampling technique to accelerate the sampling procedure. 
We provide detailed analysis of the proposed generative model in Sec.~\ref{sec:diff}, the Texture diffusion blocks in Sec.~\ref{sec:tdb} and our disentangled guidance based sampling technique in Sec.~\ref{sec:cfg}.

\subsection{Texture-Conditioned Diffusion Model}
\label{sec:diff}
The generative modeling scheme of PIDM is based on the Denoising diffusion probabilistic model~\cite{ho2020denoising} (DDPM). The general idea of DDPM is to design a diffusion process that gradually adds noise to the data sampled from the target distribution $\bm{y}_0 \sim q(\bm{y}_0)$, while the backward denoising process attempts to learn the reverse mapping. The denoising diffusion process eventually converts an isotropic Gaussian noise $\bm{y}_T \sim \mathcal{N}(0,\mathbf{I})$ into the target data distribution in $T$ steps. Essentially, this scheme divides a complex distribution-modeling problem into a set of simple denoising problems. 
The forward diffusion path of DDPM is a Markov chain with the following conditional distribution: 
\begin{equation}
    q(\bm{y}_t|\bm{y}_{t-1}) = \mathcal{N}(\bm{y}_t;\sqrt{1-\beta_t}\bm{y}_{t-1}, \beta_{t}\mathbf{I}).
\end{equation}
where $t \sim [1,T]$ and $\beta_1, \beta_2,...,\beta_T$ is a fixed variance schedule
with $\beta_t \in (0,1)$. Using the notation $\alpha_t=1-\beta_t$ and $\Bar{\alpha_t} = \prod_{i=1}^{t}\alpha_i$, we can sample from $q(\bm{y}_t|\bm{y}_0)$ in a closed form at an arbitrary timestep $t$: $
    \bm{y}_t = \sqrt{\Bar{\alpha_t}}\bm{y}_0 + \sqrt{1-\Bar{\alpha_t}}\epsilon$,  where $\epsilon \sim \mathcal{N}(0,\mathbf{I})$.
The true posterior $q(\bm{y}_{t-1}|\bm{y_t})$ can be approximated by a deep neural network to predict the mean and variance of $\bm{y}_{t-1}$ with the following parameterization,
\begin{equation}
\begin{split}
    p_{\theta}(\bm{y}_{t-1}|\bm{y}_{t}, \bm{x}_p, \bm{x}_s) = \mathcal{N}(& \bm{y}_{t-1};
    \mu_\theta(\bm{y}_{t}, t, \bm{x}_p, \bm{x}_s), \\
    & \Sigma_\theta(\bm{y}_{t}, t, \bm{x}_p, \bm{x}_s)).
\end{split}
\end{equation}
\textbf{Noise prediction module $\mathcal{H}_N$:} Instead of directly deriving $\mu_\theta$ following~\cite{ho2020denoising}, we predict the noise $\bm{\epsilon}_\theta(\bm{y}_t, t, \bm{x}_p, \bm{x}_s)$ using our noise prediction module $\mathcal{H}_N$. The noisy image $\bm{y}_t$ is concatenated with the target pose $\bm{x}_p$ and passed through $\mathcal{H}_N$ to predict the noise. $\bm{x}_p$ will guide the denoising process and ensure that the intermediate noise representations and the final image follow the given skeleton structure. To inject the desired texture patterns into the noise predictor branch, we provide the multiscale features of the texture encoder $\mathcal{H}_E$ through \textit{Texture diffusion blocks} (TDB).
To train the denoising process, we first generate a noisy sample $\bm{y}_t \sim q(\bm{y}_t|\bm{y}_0)$ by adding Gaussian noise $\epsilon$ to $\bm{y}_0$, then train a conditional denoising model $\bm{\epsilon}_\theta(\bm{y}_t, t, \bm{x}_p, \bm{x}_s)$ to predict the added noise using a standard MSE loss:
\begin{equation}
    L_{\text{mse}}=\mathbb{E}_{t\sim[1,T],\bm{y}_0\sim q(\bm{y_0}), \epsilon} \left \| \epsilon-\bm{\epsilon}_\theta(\bm{y}_t, t, \bm{x}_p, \bm{x}_s)\right \|^2.
\end{equation}
Nichol \etal~\cite{nichol2021improved} present an effective learning strategy as an improved version of DDPM with fewer steps needed and applies an additional loss term $L_{\text{vib}}$ to learn the variance $\Sigma_\theta$. The overall hybrid objective that we adopt is as follows:
\begin{equation}
    L_{\text{hybrid}} = L_{\text{mse}} + L_{\text{vib}}.
\end{equation}

\subsection{Texture Diffusion Blocks (TDB)}
\label{sec:tdb}
To mix the style of the source image within the noise prediction branch, we employ cross-attention based TDB units that are embedded in different layers of $\mathcal{H}_N$. Let $\bm{F}_h^l$ be the noise features in layer $l$ of the noise prediction branch. Given the multiscale texture features $\bm{F}_s$ derived from the $\mathcal{H}_E$ as input to TDB units, the attention module essentially computes the region of interest with respect to each query position, which is important to subsequently denoise the given noisy sample in the direction of the desired texture patterns. The keys $\bm{K}$ and values $\bm{V}$ are derived from $\mathcal{H}_E$ while queries $\bm{Q}$ are obtained from noise features $\bm{F}_h^l$. The attention operation is formulated as follows:
\begin{equation}
\begin{split}
    \bm{Q} = \phi^l_q(\bm{F}_h^l), \quad \bm{K} = \phi^l_k(\bm{F}_s), \quad \bm{V} = \phi^l_v(\bm{F}_s)\\
    \bm{F}_{att}^{l} = \frac{\bm{Q}\bm{K}^T}{\sqrt{C}}, \quad \bm{F_{o}^{l}} = \bm{W^l}\text{softmax}(\bm{F}_{att}^{l})\bm{V}+\bm{F}_h^l,
\end{split}
\end{equation}
where $\bm{\phi^l_q}$, $\bm{\phi^l_k}$, $\bm{\phi^l_v}$ are layer-specific $1\times1$ convolution operators. $\bm{W^l}$ refers to learnable weights to generate final cross-attended features $\bm{F_{o}^{l}}$. We adopt TDB for the feature at specific resolutions, \ie, $32 \times 32$, $16 \times 16$, and $8 \times 8$.

\begin{figure*}[t!]
\begin{center}
   \includegraphics[width=1\textwidth]{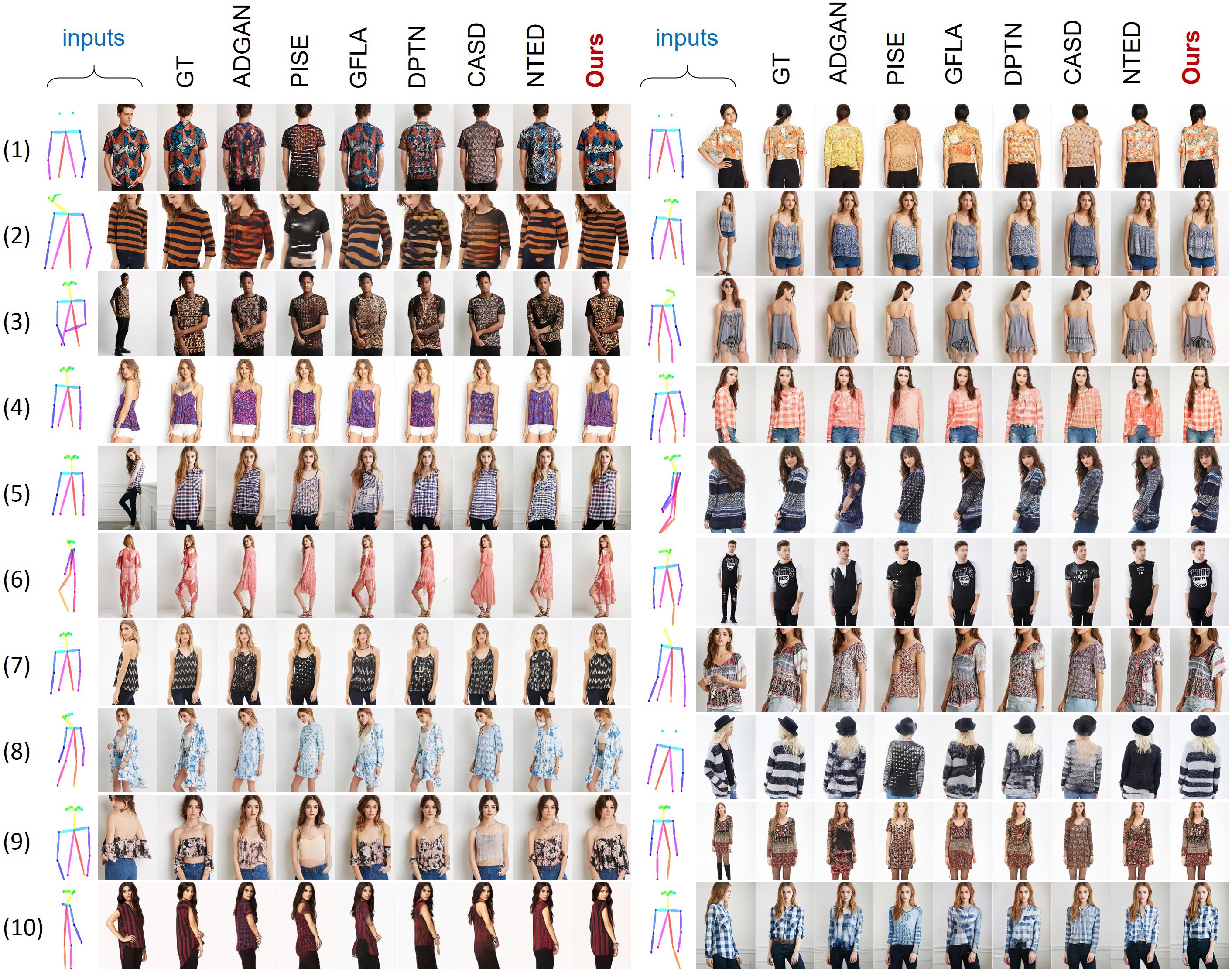}\vspace{-0.5cm}
\end{center}
\caption{Qualitative comparisons with several state-of-the-art models on the DeepFashion dataset. The inputs to the model are the target pose $\bm{x}_p$ and the source image $\bm{x}_s$. From left-to-right the results are of  ADGAN~\cite{men2020controllable}, PISE~\cite{zhang2021pise}, GFLA~\cite{ren2020deep}, DPTN~\cite{Zhang_2022_CVPR}, CASD~\cite{zhou2022cross}, NTED~\cite{ren2022neural} and ours respectively. \emph{(figure best viewed in zoom)}
}
\label{fig:qual}
 \vspace{-0.4cm}
\end{figure*}
\subsection{Disentangled Guidance based Sampling}
\label{sec:cfg}

Once the model learns the conditional distribution, inference is performed by first sampling a Gaussian noise $\bm{y}_T\sim\mathcal{N}(0, \bm{I})$ and then sampling from $p_{\theta}(\bm{y}_{t-1}|\bm{y}_{t}, \bm{x}_p, \bm{x}_s)$, from $t = T$ to $t = 1$ in an iterative manner. While the generated images using the vanilla sampling technique look photo-realistic, they often do not strongly correlate with the conditional source image and target pose input. To amplify the effect of the conditioning signal $\bm{x}_s$ and $\bm{x}_p$ in the sampled images, we adapt Classifier-free guidance~\cite{ho2022classifier} in our multi-conditional sampling procedure. We observe that in order to sample images that
not only fulfil the style requirement, but also ensure perfect alignment with the target pose input, it is important to employ \textit{disentangled guidance} with respect to both style and pose. To enable disentangled guidance, we use the following equation:
\begin{equation}
    \bm{\epsilon}_{\text{cond}} = \bm{\epsilon}_{\text{uncond}} + w_p\bm{\epsilon}_{\text{pose}} + w_s\bm{\epsilon}_{\text{style}},
\end{equation}
where $\epsilon_{\text{uncond}}=\epsilon_{\theta}(\bm{y}_t,t,\emptyset,\emptyset)$ is the unconditioned prediction of the model, where we replace both conditions with the all-zeros tensor $\emptyset$. The pose-guided prediction and the style-guided prediction are respectively represented by $\epsilon_{\text{pose}}=\epsilon_{\theta}(\bm{y}_t,t,\bm{x}_p,\emptyset)-\epsilon_{\text{uncond}}$ and $\epsilon_{\text{style}}=\epsilon_{\theta}(\bm{y}_t,t,\emptyset,\bm{x}_s)-\epsilon_{\text{uncond}}$. $w_p$ and $w_s$ are guidance scale corresponding to pose and style. In practice, the diffusion model learns both conditioned and unconditioned distributions during training by randomly setting conditional variables $\bm{x}_p$ and $\bm{x}_s = \emptyset$ for $\eta$\% of the samples, so that $\epsilon_{\theta}(\bm{y}_t,t,\emptyset,\emptyset)$ approximates $p(\bm{y}_0)$ more faithfully.

\section{Experiments}

\noindent\textbf{Datasets:} We carry out experiments on DeepFashion In-shop Clothes Retrieval Benchmark~\cite{liu2016deepfashion} and Market-1501~\cite{zheng2015scalable} dataset. DeepFashion contains 52,712 high-resolution images of fashion models. Following the same data configuration in~\cite{zhu2019progressive}, we split this dataset into training and testing subsets with 101,966 and 8,570 pairs, respectively. Skeletons are extracted by OpenPose~\cite{cao2017realtime}. Market-1501 contains 32,668 low-resolution images. The images vary in terms of the viewpoints, background, illumination, \etc. For both datasets, personal identities of the training and testing sets do not overlap. 

\noindent\textbf{Evaluation Metrics:} We evaluate the model using three different metrics. \textit{Structure Similarity Index Measure} (SSIM)~\cite{wang2004image} and \textit{Learned Perceptual Image Patch Similarity} (LPIPS)~\cite{zhang2018unreasonable} are used to quantify the reconstruction accuracy. SSIM calculates the pixel-level image similarity, while LPIPS computes the distance between the generated images and reference images at the perceptual domain. \textit{Fr\`echet Inception Distance} (FID)~\cite{heusel2017gans} is used to measure the realism of the generated images. It calculates the Wasserstein-2 distance between distributions of the generated images and the ground-truth images.

\noindent\textbf{Implementation Details:} Our PIDM model has been trained with $T = 1000$ noising steps and a linear noise schedule. During training, we adopt an exponential moving average (EMA) of the denoising network weights with 0.9999 decay. In all experiments, we use a batch size of 8. Adam optimizer is used with learning rate set to $2e^{-5}$. For disentangled guidance, we use $\eta=10$. For sampling, the values of $w_p$ and $w_s$ are set to $2.0$. For the DeepFashion dataset, we train our model using $256 \times 176$ and $512 \times 352$ images. For Market-1501, we use $128 \times 64$ images.

\begin{table}
\begin{center}
\caption{Quantitative Comparison of the proposed PIDM with several state-of-the-art models in terms of \textit{Fr\`echet Inception Distance} (FID), \textit{Structure Similarity Index Measure} (SSIM) and \textit{Learned Perceptual Image Patch Similarity} (LPIPS). The results are shown on both $256 \times 176$ and $512\times 352$
resolution for DeepFashion and $128 \times  64$ resolution for Market-1501 dataset. 
}
\label{tab:main_table}
\setlength{\tabcolsep}{5pt}
\scalebox{0.8}{
\begin{tabular}{l|l|c|c|c}
\toprule[0.4mm]
\rowcolor{mygray} 
\cellcolor{mygray}Dataset & Methods &
  FID($\downarrow$) &  SSIM($\uparrow$) & LPIPS($\downarrow$)\\ \midrule
    \multirow{10}*{\begin{tabular}{l}DeepFashion~\cite{liu2016deepfashion}\\
                                    ($256\times176$)\\
                \end{tabular}}  
                
& Def-GAN \cite{siarohin2018deformable}    &18.457 &0.6786 &0.2330 \\  
& PATN \cite{zhu2019progressive}   &20.751 &0.6709  &0.2562\\
& ADGAN \cite{men2020controllable}   &14.458 &0.6721 &0.2283  \\
& PISE \cite{zhang2021pise}      &13.610 &0.6629 &0.2059 \\
& GFLA \cite{ren2020deep}      &10.573 &0.7074 &0.2341\\
& DPTN \cite{Zhang_2022_CVPR}      &11.387 &0.7112 &0.1931\\\
& CASD \cite{zhou2022cross}      &11.373 &0.7248 &0.1936\\
& NTED \cite{ren2022neural}      &8.6838 &0.7182 &0.1752\\
& \textbf{PIDM (Ours)}            &\textbf{6.3671} &\textbf{0.7312} &\textbf{0.1678} \\ \midrule
\multirow{3}*{\begin{tabular}{l}DeepFashion~\cite{liu2016deepfashion}\\
                                    ($512\times352$)\\
                \end{tabular}} 
& CocosNet2 \cite{Zhou_2021_CVPR}     &13.325 &0.7236 &0.2265  \\
& NTED \cite{ren2022neural}      &7.7821 &0.7376 &0.1980 \\

& \textbf{PIDM (Ours)}          &\textbf{5.8365} &\textbf{0.7419} &\textbf{0.1768} \\
\midrule
\multirow{4}*{\begin{tabular}{l}Market-1501~\cite{zheng2015scalable}\\
                                    ($128\times64$)\\
                \end{tabular}} 
& Def-GAN \cite{siarohin2018deformable}     &25.364 &0.2683 &0.2994  \\
& PTN \cite{zhu2019progressive}      &22.657 &0.2821 &0.3196 \\
& GFLA \cite{ren2020deep}      &19.751 &0.2883 &0.2817 \\
& DPTN \cite{Zhang_2022_CVPR} &18.995 &0.2854 &0.2711\\
& \textbf{PIDM (Ours)}            &\textbf{14.451} &\textbf{0.3054} & \textbf{0.2415} \\
\bottomrule[0.4mm]
\end{tabular}
}
\end{center}
\end{table}

\subsection{Quantitative and Qualitative Comparisons}

We quantitatively compare (Tab.~\ref{tab:main_table}) our proposed PIDM with several state-of-the-art methods, including Def-GAN \cite{siarohin2018deformable}, PATN \cite{zhu2019progressive}, ADGAN~\cite{men2020controllable}, PISE~\cite{zhang2021pise}, GFLA~\cite{ren2020deep}, DPTN~\cite{Zhang_2022_CVPR}, CASD~\cite{zhou2022cross}, CocosNet2 \cite{Zhou_2021_CVPR} and NTED~\cite{ren2022neural}. The experiments are done on both $256 \times 176$ and $512\times 352$ resolution for DeepFashion and $128 \times  64$ resolution for Market-1501 dataset.  Tab.~\ref{tab:main_table} shows that our model achieves the best FID score indicating that our model can generate higher-quality images compared to the previous approaches. Furthermore, PIDM
performs favorably against other methods in terms of the reconstruction metrics SSIM and LPIPS. This means that our model can generate images with not only accurate structures, but also can correctly transfer the texture of the source image to the target pose.

\begin{figure}[t!]
\begin{center}
   \includegraphics[width=1\linewidth]{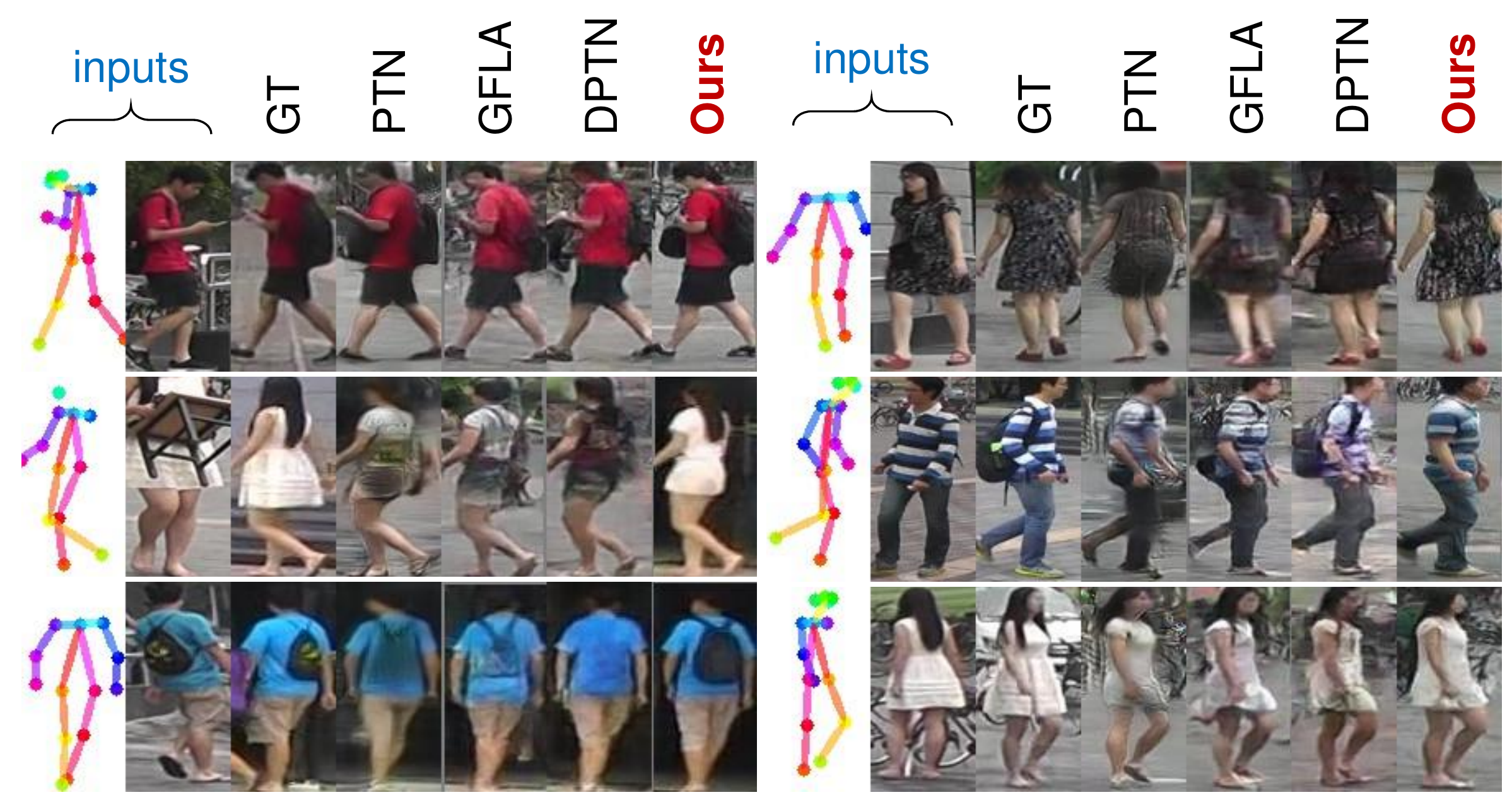}\vspace{-0.5cm}
\end{center}
\caption{Qualitative comparisons with several state-of-the-art models on the Market-1501 dataset. From left-to-right the results are of  PTN~\cite{zhu2019progressive}, GFLA~\cite{ren2020deep}, DPTN~\cite{Zhang_2022_CVPR} and ours respectively. 
}
\label{fig:qual_market}
 \vspace{-0.4cm}
\end{figure}

In Fig.~\ref{fig:qual}, we present a comprehensive visual comparison of our method with other state-of-the-art frameworks on DeepFashion dataset\footnote{\label{note1}Additional results are provided in supplementary material.}. The results of the baselines are obtained using pre-trained models provided by the corresponding authors. It can be observed that PISE~\cite{zhang2021pise} and ADGAN~\cite{men2020controllable} fail to generate sharp images and cannot keep the consistency of shape and texture. While GFLA~\cite{ren2020deep} somewhat preserves the texture in the source image, it struggles to obtain reasonable results for the invisible regions of the source image (\eg, 4\(\sim\)5-$th$ rows in the left column). NTED~\cite{ren2022neural} and CASD~\cite{zhou2022cross} improve the results slightly but they are still not able to adequately preserve the source appearance in complex scenarios (\eg, 1\(\sim\)3-$th$ rows in the left and 7\(\sim\)10-$th$ rows in the right column). In comparison, our proposed PIDM accurately retains the appearance of the source while also producing images that are more natural and sharper. Moreover, even if the target pose is complex (\eg, 10-$th$ row in the left column), our method can still generate it precisely. We also visually compare our method with other baselines on the Market-1501 dataset in Fig.~\ref{fig:qual_market}. Even with the complex background, PIDM still performs favorably in generating photo-realistic results as well as a consistent appearance with the source image.

\begin{figure}[t!]
\begin{center}
   \includegraphics[width=0.8\linewidth]{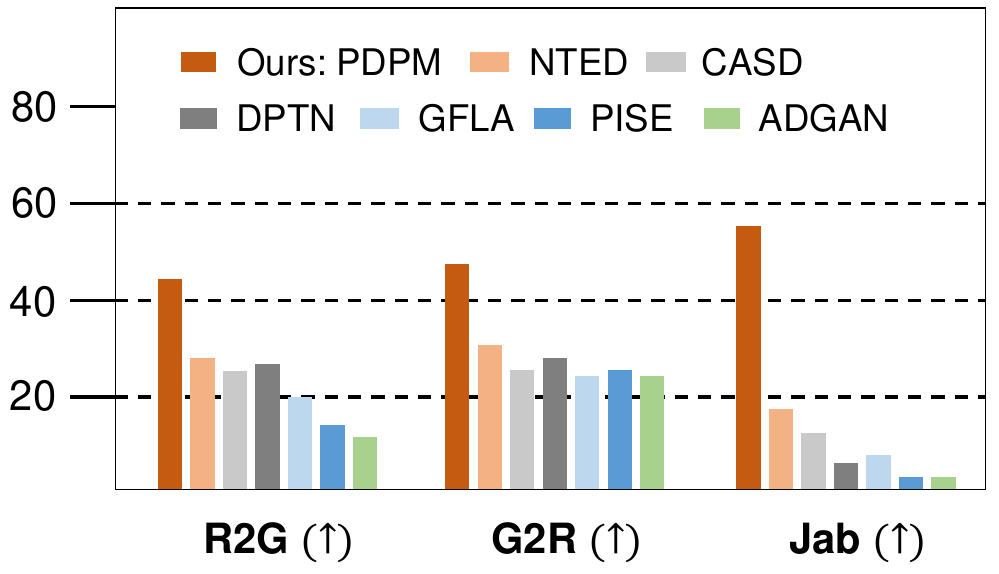}\vspace{-0.5cm}
\end{center}
\caption{User study results on DeepFashion dataset in terms of \textit{R2G}, \textit{G2R} and \textit{Jab} metric. Higher
values indicate PIDM is preferred more often over the compared approaches.
}
\label{fig:user}
 \vspace{-0.4cm}
\end{figure}

\subsection{User Study}
To demonstrate the effectiveness of our model in terms of human perception, we present our user study on 100 human participants. The user study is conducted to measure different aspects of our model with respect to ground truth images and generated images using other methods. \textbf{(i)} To compare with ground-truth images, we randomly select 30 generated images and 30 real images from the test set. Participants are required to determine whether a given image is real or fake. Following~\cite{zhu2019progressive, zhou2022cross}, we adopt two metrics: \textit{R2G} and \textit{G2R}. \textit{R2G} is the percentage of the real images classified as generated images and \textit{G2R} is the percentage of the generated images classified as real images. 
\textbf{(ii)} To compare with other methods, we randomly select 30 sets of images where each set includes source image, target pose, ground-truth and images generated by our method and the baseline. The participants are required to select the best image with respect to the provided source image and ground truth. The participants are advised to provide their response based on the ability of each competing approach towards producing accurate texture patterns and pose structure. We quantify this using another metric called \textit{Jab} which is defined as the percentage of images considered the best among all models. Higher values of these three metrics mean better performance. The result of the study is shown in Fig.~\ref{fig:user}. PIDM performs favorably against all baselines for all three metrics on 
DeepFashion dataset. For instance, PIDM images were interpreted as real images 48\% (\textit{G2R}) out of total cases, which is nearly 18\% higher than the second best model. Our \textit{Jab} score is 56\% showing that the participants favor our approach more frequently than other methods.

\begin{table}[t!]
\begin{center}
\caption{Impact of our texture diffusion blocks (TDB) and distangled classifier-free (DCF) guidance in our proposed diffusion model on DeepFashion dataset. }\vspace{-0.3cm}
\label{tab:ablation2}
\setlength{\tabcolsep}{3pt}
\scalebox{0.75}{
\begin{tabular}{l|c|c|c}
\toprule[0.4mm]

\rowcolor{mygray}
Methods & FID($\downarrow$) & SSIM($\uparrow$) & LPIPS($\downarrow$) \\ \midrule

B1: \texttt{Baseline${}^\dagger$} (concat)        &10.813 &0.6911 &0.2112 \\
B2: \texttt{Baseline}         &9.8510 &0.7005 &0.1983\\
B3: \texttt{Baseline + TDB}        &7.5133 &0.7178 &0.1870\\
B4: \texttt{Baseline + TDB + CF-guidance}      &6.8176 &0.7195 &0.1769\\
\midrule
\textbf{Ours: Baseline + TDB + DCF-guidance}     
     &\textbf{6.3671}  &\textbf{0.7312} &\textbf{0.1676}\\  \bottomrule[0.4mm]
\end{tabular}
}
\end{center}\vspace{-0.7cm}
\end{table}

\begin{figure}[t!]
\begin{center}
   \includegraphics[width=1\linewidth]{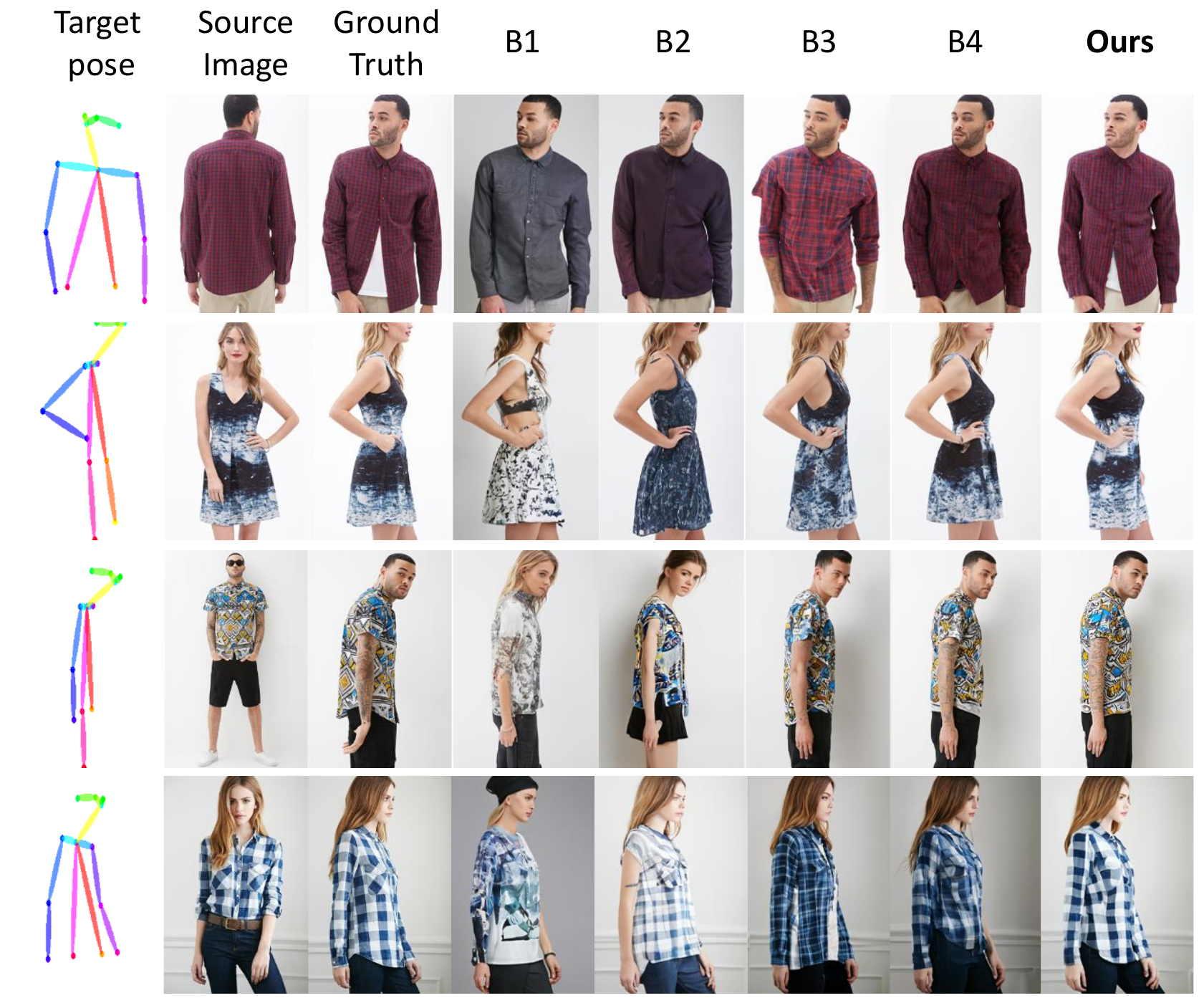}\vspace{-0.5cm}
\end{center}
\caption{Qualitative ablation results on the DeepFashion dataset. The images in this figure correspond to the ablation studies in Tab.~\ref{tab:ablation2}. Compared to the B1 (\texttt{Baseline${}^\dagger$}), the B2 (\texttt{Baseline}) benefits from an additional texture encoder improving the correlation with the source appearance. The B3 (\texttt{Baseline+TDB}) further improves the results using texture diffusion module (TDB) %integrated into the noise prediction module 
that effectively model the complex interplay between appearance and pose information. Our final
proposed PIDM referred as \texttt{Baseline+TDB+DCF-guidance} employs the disentangled classifer-free (DCF) guidance. Contrary to the B4 (\texttt{Baseline+TDB+CF-guidance}), our proposed PIDM employs \textit{disentangled guidance} with respect to both style and pose in order to tightly align the output image style and pose with the source image appearance and target pose. }
\label{fig:ablation}
 \vspace{-0.4cm}
\end{figure}

\begin{figure*}[t!]
\begin{center}
   \includegraphics[width=1\linewidth]{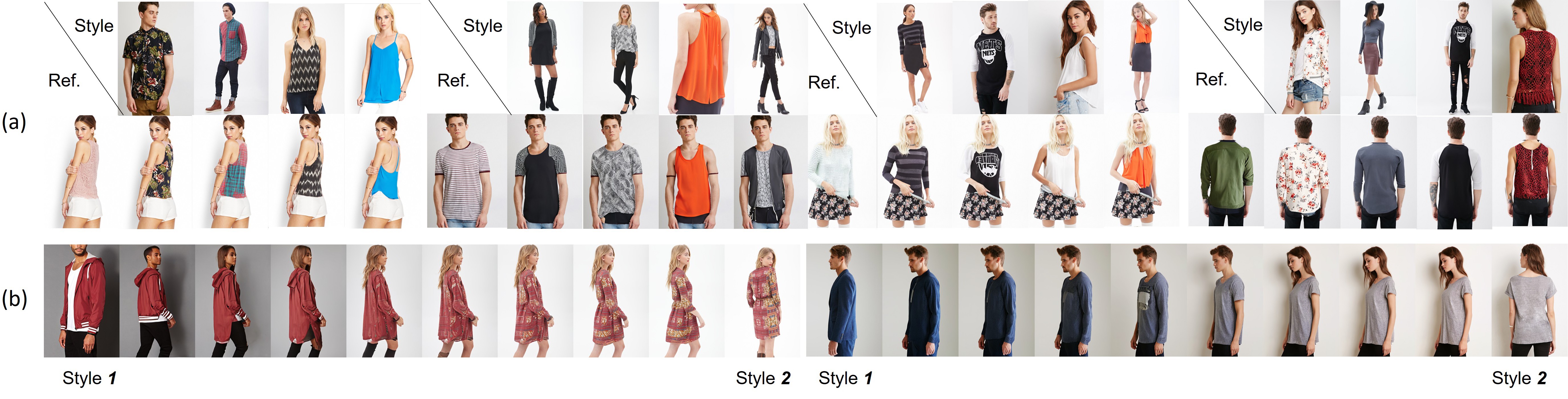}\vspace{-0.5cm}
\end{center}\vspace{-0.2cm}
\caption{(a) Qualitative evaluation of our proposed PIDM for appearance control. The results demonstrate the seamless editing capabilities of our model. Images are generated by controlling the appearance of the reference image while maintaining the person's pose and identity. For each example, the first row contains the style images and the second row contains the generated images. (b) The interpolation results show that the clothes' texture gradually changes from the style of the left image to that of the right image. \emph{(figure best viewed in zoom)}}
\label{fig:control}
 \vspace{-0.3cm}
\end{figure*}

\subsection{Ablation Study}

We perform multiple ablation studies to validate the merits of the proposed contributions. Tab.~\ref{tab:ablation2} shows the impact of texture diffusion blocks (TDB) and distangled classifier-free (DCF) guidance on the DeepFashion dataset. \texttt{Baseline${}^\dagger$} neither employs a separate texture encoder nor uses TDB units. It only comprises of a UNet-based noise prediction module where the target pose and source image are concatenated with the noisy image and passed through the module to output a denoised version of the image. We extend the \texttt{Baseline${}^\dagger$} with additional texture encoder to extract meaningful texture representations from the source image which are concatenated with respective layers of the noise prediction module. This is denoted as \texttt{Baseline} in Tab.~\ref{tab:ablation2}. While the \texttt{Baseline} is able to generate realistic person images, it has a limited ability to retain the appearance of the source image as shown in Fig.~\ref{fig:ablation}. To effectively model the complex interplay between appearance and pose information, we integrate texture
diffusion module (TDB) into the noise prediction module. We refer to this as \texttt{Baseline+TDB}. On the DeepFahsion dataset, the \texttt{Baseline} achieves an FID score of 9.8510. In comparison to the \texttt{Baseline}, the \texttt{Baseline+TDB} improves the FID by a margin of 2.3377. 
While the generated images using the vanilla sampling technique look photo-realistic, they often do not strongly correlate with the conditional source image and target pose input. To improve the correlation, we first use vanilla classifier-free (CF) guidance during the sampling. While the vanilla CF guidance slightly improves the results, we observe that in order to sample images that not only fulfil the style requirement, but also ensure perfect alignment with the target pose input, it is important to employ disentangled guidance with respect to both style and pose. We here refer to our final proposed approach \texttt{Baseline+TDB+DCF-guidance} where we employ distangled classifier-free (DCF) guidance to tightly align the output image style and pose with the source image appearance and target pose. In comparison to the \texttt{Baseline+TDB}, our proposed DCF guidance improves the FID, SSIM and LPIPS by a margin of 1.1462, 0.0134 and 0.0194 respectively.

\subsection{Appearance Control and Editing}
Our proposed PIDM inherits the flexibility and controllability of diffusion models that enable appearance control by combining cloth textures extracted from style images into the reference image. Given a source (style) image $\bm{x}_{s}$, a reference image $\bm{y}^{ref}$, and a binary mask $\bm{m}$ that marks the region of interest in the reference image, the problem is to generate an image $\bar{\bm{y}}$, s.t. the appearance of $\bar{\bm{y}}\odot \bm{m}$ is consistent with the source image $\bm{x}_{s}$, while the complementary area remains same. To achieve this, we first calculate $\bm{y}_t^{ref}$ in a time step $t$: $
\bm{y}_t^{ref} = \sqrt{\Bar{\alpha_t}}\bm{y}^{ref} + \sqrt{1-\Bar{\alpha_t}}\epsilon$, where $\epsilon \sim \mathcal{N}(0,\mathbf{I})$. During inference, starting from a Gaussian noise $\bm{y}_T\sim\mathcal{N}(0, \mathbf{I})$, we predict $\bm{y}_t$ iteratively using the trained diffusion model from $t = T$ to $t = 1$. In each step $t$, we use the binary mask $\bm{m}$ to retain the unmasked regions of $\bm{y}^{ref}$ by using the relation: $\bm{y}_{t}=\bm{m}\odot\bm{y}_{t}+(1-\bm{m})\odot\bm{y}_t^{ref}$. The results of the appearance control task are shown in Fig.~\ref{fig:control}(a). We observe that our model can seamlessly combine the areas of interest and generate coherent output images with realistic textures.% Please note we don\'t use any additional parser maps during training still our model is able to learn 

\noindent\textbf{Style Interpolation:}
In Fig.~\ref{fig:control}(b), we show our interpolation results between two style images. We use DDIM~\cite{song2020denoising} sampling to enable smooth interpolation using our trained diffusion model. Specifically, we use spherical linear interpolation between noises $\bm{y}_T^{1}$ and $\bm{y}_T^{2}$, and linear interpolation between style features $\bm{F}_s^{1}$ and $\bm{F}_s^{2}$. As shown in Fig.~\ref{fig:control}(b), the texture of the clothes gradually changes from the style of the left source image to that of the right source image. 

\subsection{Application to Person Re-identification} 
Here, we evaluate the applicability of the images generated by our PIDM as a source of data augmentation for the downstream task of person re-identification (re-ID). We perform the re-ID experiment on Market-1501 dataset. We randomly select 20\%, 40\%, 60\%, and 80\% of total training set of real Market-1501 dataset such that at least one image per identity is selected. We denote the obtained set as $\mathbb{D}_{tr}$. We first initialize a ResNet50 backbone network using $\mathbb{D}_{tr}$ for the re-ID task. We refer to this as \textit{Standard} in Tab.~\ref{tab:reid}. Then, we augment $\mathbb{D}_{tr}$
with images generated by our PIDM. The images are generated using randomly chosen image of the same identity in $\mathbb{D}_{tr}$ as the source image. The target pose is randomly selected from $\mathbb{D}_{tr}$. Consequently, an augmented training set $\mathbb{D}_{aug}$ is created from all these generated images and the real set $\mathbb{D}_{tr}$. We finetune the ResNet50 backbone using the augmentation set $\mathbb{D}_{aug}$. As shown in Tab.~\ref{tab:reid}, PIDM achieves consistent improvements over previous works.%, suggesting the proposed method can generate realistic human images and be more effective for the re-ID task.  

\begin{table}[t!]
\begin{center}
\caption{The person re-ID results on ResNet50 backbone in terms of mAP scores using images generated by different methods. The right-most column shows results when 30K generated images are added to the real set. \textit{Standard} means no augmentation is used.
}\vspace{-0.2cm}
\label{tab:reid}
\setlength{\tabcolsep}{8pt}
\scalebox{0.7}{
\begin{tabular}{l|c|c|c|c|c} 
\toprule[0.4mm]
\rowcolor{mygray}  &
  \multicolumn{4}{c|}{\cellcolor{mygray}Percentage of real images} &   \\ 
\rowcolor{mygray} 
\multirow{-2}{*}{\cellcolor{mygray}Methods} &
  20\% & 40\% & 60\% & 80\%  &

  \multirow{-2}{*}{\cellcolor{mygray}100\%(+30K)}\\ \midrule
Standard  &33.4 &56.6 &64.9 &69.2 &76.7\\   \midrule
PTN~\cite{zhu2019progressive}       &55.6 &57.3 &67.1 &72.5 &76.8  \\
GFLA~\cite{ren2020deep}     &57.3 &59.7 &67.6 &73.2 &76.8 \\
DPTN~\cite{Zhang_2022_CVPR}      &58.1 &62.6 &69.0 &74.2  &77.1 \\\midrule
\textbf{PIDM (Ours)}         &\textbf{61.3} &\textbf{64.8} &\textbf{71.6} &\textbf{75.3} &\textbf{78.4}  \\ \bottomrule[0.4mm]
\end{tabular}}
\end{center} 
\vspace{-0.3cm}
\end{table}

\section{Conclusion}

We proposed a diffusion-based approach, PIDM, for pose-guided person image generation. PIDM disintegrates the complex transfer problem into a series of simpler forward-backward denoising steps.  This helps in learning plausible source-to-target transformation trajectories that result in faithful textures and undistorted appearance details. We introduce \textit{texture diffusion module} and \textit{disentangled classifier-free guidance} to accurately model the correspondences between appearance and pose information available in source and target images. We show the effectiveness of PIDM on two datasets by performing extensive qualitative, quantitative and human-based evaluations. In addition, we show how our generated images can help in downstream tasks such as person re-identification.

%%%------------------
%% main architecture figures 

%%%%%%%%% REFERENCES
{\small
\bibliographystyle{ieee_fullname}
\bibliography{egbib}
}
\end{document}